\documentclass[10pt,twocolumn,letterpaper]{article}

\usepackage{iccv}
\usepackage{times}
\usepackage{epsfig}
\usepackage{graphicx}
\usepackage{amsmath}
\usepackage{amssymb}
\usepackage{xcolor}
\usepackage{multirow}
\usepackage{colortbl}
\usepackage{booktabs}
\usepackage{algorithm}
\usepackage{algorithmic}

\usepackage[breaklinks=true,bookmarks=false]{hyperref}

\iccvfinalcopy 

\def\iccvPaperID{****} 

\ificcvfinal\fi

\begin{document}

\title{Towards Black-box Adversarial Example Detection: A Data Reconstruction-based Method}

\author{
Yifei Gao$^{1}$, Zhiyu Lin$^{1}$, Yunfan Yang$^{1}$, Jitao Sang$^{1,2}$\\
Beijing Jiaotong University, China\\
Peng Cheng Lab, Shenzhen 518066, China\\
{\tt\small \{yifeigao, zyllin, yunfanyang, jtsang\}@bjtu.edu.cn}
}

\maketitle
\ificcvfinal\thispagestyle{empty}\fi

\begin{abstract}
  Adversarial example detection is known to be an effective adversarial defense method. Black-box attack, which is a more realistic threat and has led to various black-box adversarial training-based defense methods, however, does not attract considerable attention in adversarial example detection. In this paper, we fill this gap by positioning the problem of black-box adversarial example detection (BAD). Data analysis under the introduced BAD settings demonstrates (1) the incapability of existing detectors in addressing the black-box scenario and (2) the potential of exploring BAD solutions from a data perspective. To tackle the BAD problem, we propose a data reconstruction-based adversarial example detection method. Specifically, we use variational auto-encoder (VAE) to capture both pixel and frequency representations of normal examples. Then we use reconstruction error to detect adversarial examples. Compared with existing detection methods, the proposed method achieves substantially better detection performance in BAD, which helps promote the deployment of adversarial example detection-based defense solutions in real-world models.
\end{abstract}

\section{Introduction}
Deep neural networks (DNNs) have recently achieved and sometimes surpassed “human-level” benchmarks across a wide range of tasks, especially in visual recognition. However, recent studies \cite{szegedy2013intriguing}  have shown that DNN models are vulnerable to human-imperceptible adversarial perturbations, which causes erroneous model predictions. The existence of adversarial examples exposes a non-negligible security risk when deploying DNN models in critical application, such as autonomous driving and facial recognition system. As a more realistic threat in real-world applications, black-box attack assume that attackers have limited access to the DNN model (referred to as the victim model) or its training data. Instead, a substitute model (referred to as the threat model) is used to improve the attack efficiency.

\begin{figure}[t]
	\centering
        \includegraphics[width=0.48\textwidth]{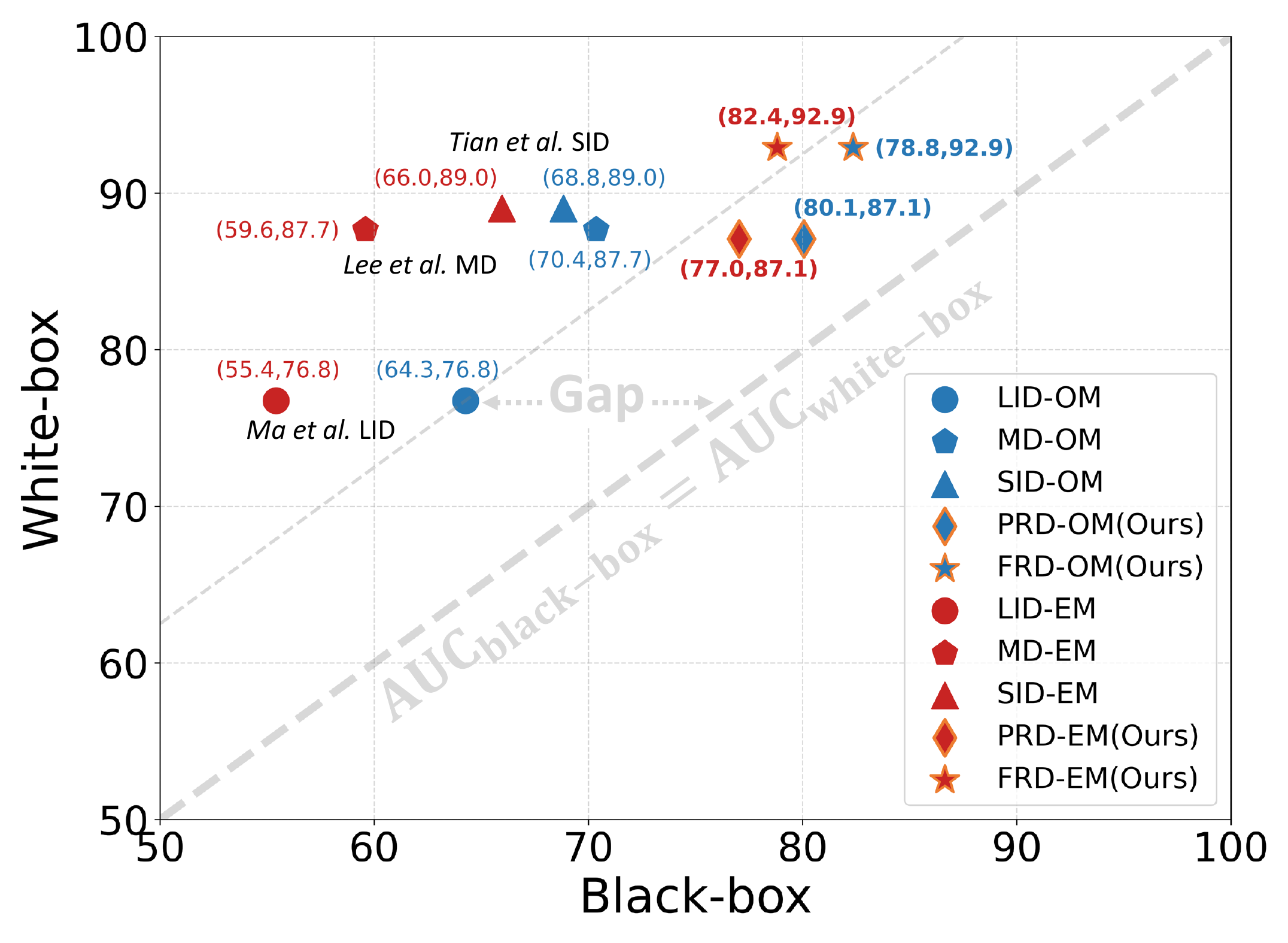}
        \caption{Detection performance on both white- and black-box adversarial example. Existing detection methods (LID, MD and SID) manifest significant performance degradation under black-box attacks detection (reflects to the gap to the diagonal). PRD and FRD are our proposed data reconstruction-based methods, which narrow the gap between white- and black-box attacks detection. SM and EM denote different black-box settings.} 
 	\label{fig:introduction}
\vspace{-0.6cm}
\end{figure}

In the attack-defense game \cite{huang2021sequential}, attackers usually hold the initiative, thus requiring the defender ought to have comparable competitiveness. Current defensive techniques can be categorized into two major groups \cite{aldahdooh2022adversarial}: (1) robust defense, such as adversarial training, which aims at improving the robustness of the DNN model and defending against both white- and black-box attacks \cite{tramer2017ensemble}; and (2) adversarial example detection, which aims to identify and reject the potential adversarial examples without modifying the victim model. Typical detection methods include LID \cite{DBLP:conf/iclr/Ma0WEWSSHB18}, MD \cite{lee2018simple} and SID \cite{tian2021detecting}. However, existing detection methods assume that attackers have full access to the victim model (e.g., threat model=victim model), making them effective only against the white-box attacks. This raises concerns about the capabilities of existing detection methods when detecting black-box adversarial examples.

To verify concerns, we first position the problem of black-box adversarial example detection (BAD) by including the black-box adversarial examples into evaluation. We re-evaluate the performance of some typical detection methods under two provided black-box settings: (1) single model (SM) attack and (2) ensemble model (EM) attack. Figure \ref{fig:introduction} illustrates the detection performance in the white-box setting (y-axis) and two black-box settings (x-axis). All existing methods manifest significant performance degradation of more than $12.5\%$, as evidence by the gaps to the diagonal (\emph{$AUC_{black-box}=AUC_{white-box}$}) in Figure \ref{fig:introduction}. This indicates an obvious insufficiency of existing methods in black-box adversarial example detection. We attribute this performance degradation to the reliance on the victim model (detailed in Sec. \ref{Sec: Section2}), which inspires us to design an effective detection method from a data-centric perspective.

In this paper, we propose a straightforward yet effective data reconstruction-based detection method, which consists of three modules: (1) Data reconstruction module, where we use a VAE \cite{kingma2013auto} approach to learn the visual representations from both pixel and frequency domains. (2) Feature extraction module, where we extract the layer activation of the images before and after reconstruction from the victim model. (3) Adversarial example detection module, where we calculate the reconstruction error and apply it as the detector features to identify the adversarial examples. We consider two versions of our method based on the choice of feature extractor. The base version uses the victim model, while the online version replaces it with a pre-trained model to further alleviate the reliance on the victim model.

Our contributions are summarized as follows: (1) We position the problem of black-box adversarial example detection (BAD) and validate the incapability of existing detection methods in addressing BAD. (2) We propose a data reconstruction-based method for adversarial example detection, which can identify the black-box attacks via reconstructing the pixel and frequency representations. (3) Extensive experiments and analyses demonstrate the effectiveness of our proposed method, which achieves the state-of-the-art performance under both white- and black-box attacks detection.

\begin{figure*}[h]
	\centering
	\includegraphics[width=1.0\textwidth]{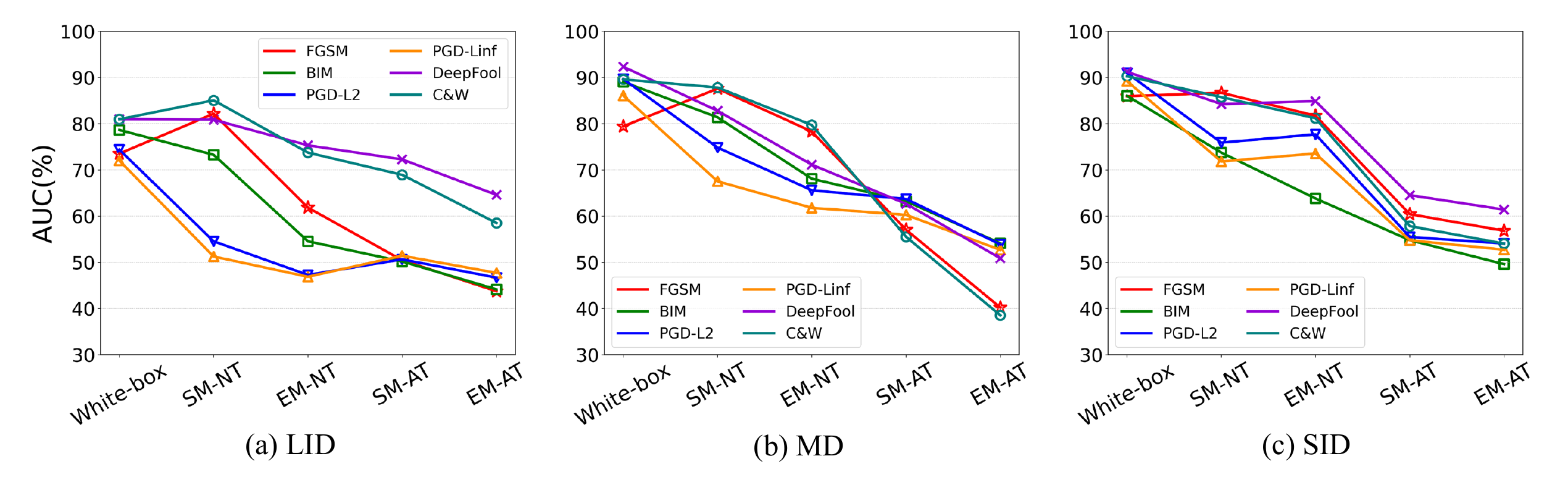}
    \vspace{-0.2cm}
    \caption{BAD evaluation on existing detection methods. We observe a performance degradation on most of the black-box settings.} 
 	\label{fig:data-snslysis}
\end{figure*}
\section{Related Works}\label{Sec: Related}
\noindent \textbf{Adversarial attack}: The Fast Gradient Sign Method (FGSM) \cite{0Explaining} is an efficient and fast attack method that performs a one-step attack along the direction orthogonal to the decision boundary. To improve the attack efficiency, the Basic Iterative Method (BIM) \cite{10.1007/978-3-030-62460-6_51} is designed by performing an iterative update using FGSM. Similarly, Projected Gradient Descent (PGD) \cite{madry2018towards}, which is the strongest first-order attack method, further improve the attack efficiency and is often applied in adversarial-training. DeepFool \cite{moosavi2016deepfool} is a powerful attack method that utilizes the geometric properties of the model to craft adversarial examples, while C\&W \cite{7958570} is an optimization-based attack. However, the above five white-box attacks are often used to evaluate model robustness, rather than considered as practical attack methods \cite{chen2017zoo}. In contrast, black-box attack methods are mainly divided into two categories: transfer-based attack and query-based attack. In transfer-based attack, adversarial examples are crafted via a substitute model \cite{huang2022transferable}. In query-based attacks, the attacker does not require full knowledge of the victim model but continuously queries the model outputs to craft adversarial examples \cite{feng2022boosting}.

\vspace{0.15cm}

\noindent\textbf{Adversarial example detection}: Most of the existing detection methods are only capable of detecting white-box examples and are heavily dependent on the layer activation of the victim model. \cite{DBLP:journals/corr/FeinmanCSG17} proposes a logistic regression detector that utilizes kernel density (KD) and Bayesian uncertainty (BU) features. However, \cite{DBLP:conf/iclr/Ma0WEWSSHB18} pointed out the limitations of KD and BU features in detecting local adversarial examples and introduced a new detection metric based on Local Intrinsic Dimensionality (LID). Alternatively, \cite{lee2018simple} uses Mahalanobis distance (MD) to detect the adversarial examples. Furthermore, \cite{tian2021detecting} proposes a Sensitivity Inconsistency Detector (SID) that employs a dual classifier with a transformed decision boundary. To address BAD problem, this paper focuses primarily on a data reconstruction-based detection method. Although previous works such as CD-VAE \cite{NEURIPS2021_CDVAE} have considered reconstruction-based methods and MagNet \cite{meng2017magnet} uses reconstruction error to detect the adversarial examples, none of them have specifically focused on black-box adversarial example detection.

\begin{table}[!tbp]
    \centering
        \caption{The black-box settings in BAD. We consider single model (SM) attack and ensemble model (EM) attack. For network architecture, we consider VGG16 \cite{vgg16}, ResNet18 \cite{resnet} and WideResNet28 (WRN28) \cite{wide}. For learning strategy, we consider adversarial training (AT) and natural training (NT). \& means model ensemble.}
        \vspace{0.15cm}
        \resizebox{0.45\textwidth}{!}{
        \setlength\tabcolsep{13pt}
        \scalebox{0.8}{
        \begin{tabular}{ccc}
            
            \toprule

             ~ & Single Model (SM) & Ensemble Model (EM) \\
            
            \midrule
            NT & VGG16/WRN28  & VGG16 \& WRN28\\
            AT & ResNet18 & VGG16 \& WRN28\\ 
            \bottomrule
            
        \end{tabular}
        }
    }
\label{tab1:black-setting}
\vspace{-0.35cm}
\end{table}

\section{Data Analysis and Motivation Justification}\label{Sec: Section2}
In this section, we formally define the problem of black-box adversarial example detection (BAD) (Sec. \ref{Sec2.1}). We provide two black-box settings and re-evaluate the performance of existing detectors under BAD (Sec. \ref{Sec2.2}). We further analyze between normal and adversarial examples, which inspires us to explore BAD solutions from a data perspective (Sec. \ref{Sec2.3}).

\subsection{Black-box Adversarial Example Detection (BAD): Definition and Settings}\label{Sec2.1}
\noindent\textbf{Definition}: Existing detection methods typically focus on the generalization of adversarial example detection from one seen attack to other unseen attacks ~\cite{tian2021detecting,cohen2020detecting,lee2018simple}. Nevertheless, those unseen adversarial examples are all generated by the victim model and do not meet the definition of black-box adversarial examples. In other words, existing detection methods are still limited to defending against the white-box attacks. We believe that including the black-box adversarial examples in the evaluation of detection methods can better reflect the real performance. In this view, we formally define BAD with the following two characteristics: 1) in the training phase, defenders use white-box adversarial examples (from the victim model) to train the detector; 2) in the testing phase, defenders evaluate the detector through black-box adversarial examples (from the threat model that is different from the victim model).

\vspace{0.15cm} 

\noindent\textbf{BAD settings}: We provide quantitative measurements for existing detection methods under BAD by introducing two settings for black-box adversarial example generation: 1) signal model (SM) attack and 2) ensemble model (EM) attack. To fully explore the influence of different threat models, we consider two black-box variables, the network architecture and the learning strategy. The detailed settings are shown in Table \ref{tab1:black-setting}. We re-evaluate the performance of LID \cite{DBLP:conf/iclr/Ma0WEWSSHB18}, MD \cite{lee2018simple} and SID \cite{tian2021detecting} methods on CIFAR-10 and CIFAR-100 dataset \cite{krizhevsky2009learning} (Results of CIFAR-100 are available in appendix and the key observations are consistent). Each detection method contains six detectors trained on FGSM \cite{0Explaining}, BIM\cite{10.1007/978-3-030-62460-6_51}, PGD ($l_{2}$ and $l_{\infty}$) \cite{madry2018towards}, DeepFool \cite{moosavi2016deepfool} and C\&W \cite{7958570} examples, respectively. Unless otherwise specified, we set ResNet18 (NT) as the default victim model.

\subsection{Performance Degradation on BAD}\label{Sec2.2}
\noindent\textbf{Observations of BAD evaluation}: Figure \ref{fig:data-snslysis} presents the detection performance (AUC scores on the y-axis) evaluated in various black-box settings (x-axis), as well as the performance in the white-box setting, where the victim model provides the adversarial examples for evaluation. The key observations are as follows: (1) All detectors (represented by line color) manifest performance degradation in black-box settings, indicating that the existing detection methods cannot well transfer to BAD. (2) Detector performance shows larger variance in the black-box settings than in the white-box setting, e.g., in the LID method, the performance difference under SM(NT) setting is much greater than that in the white-box setting. An intriguing observation is that, for the LID method, the performance of the FGSM detector in the SM(NT) is better than in the white-box setting. Moreover, FGSM detector achieves the second-best performance in SM(NT), surpassing its rank in the white-box setting ($4^{th}$ place). This reflects the unexposed detection capability of the FGSM detector in the white-box setting.

\vspace{0.15cm}

\begin{figure}[t]
	\centering
	\includegraphics[width=0.48\textwidth]{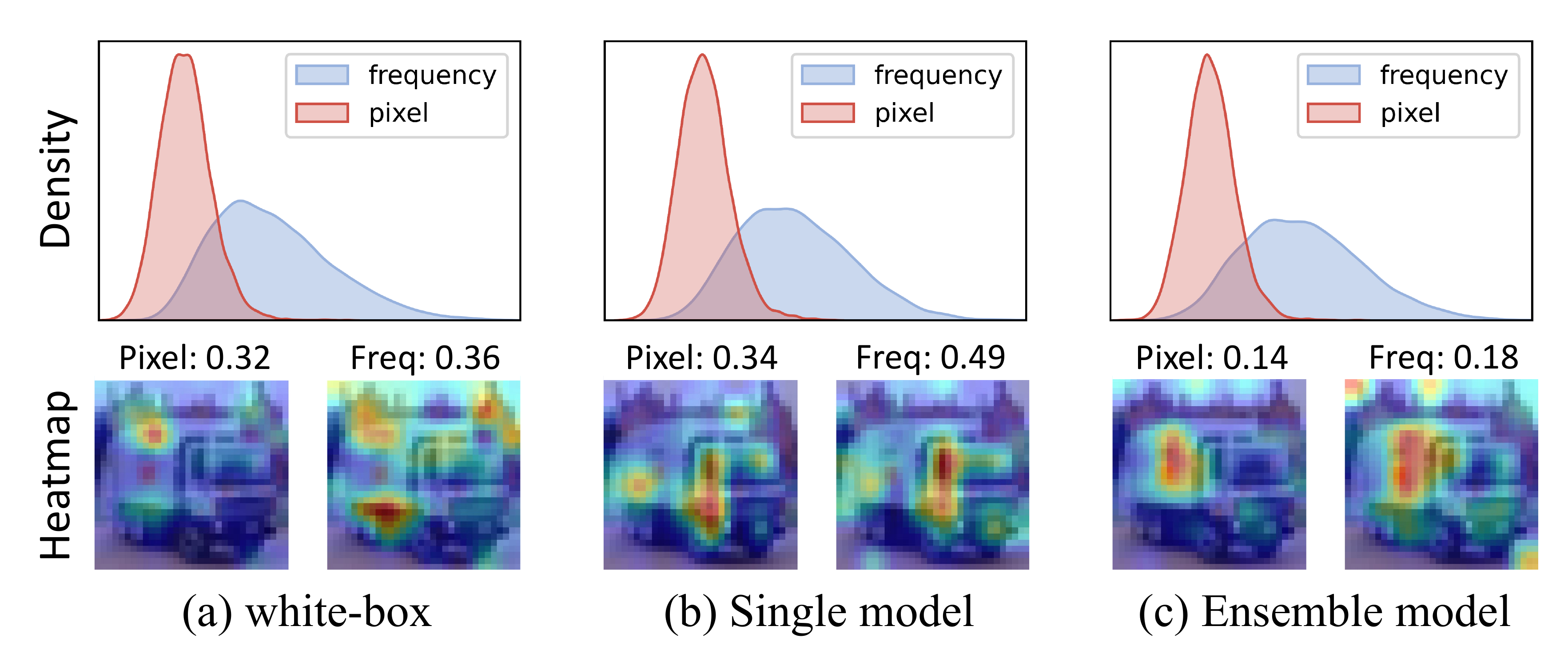}
        \caption{Difference measurements between normal and adversarial examples. The first row shows the Kernel Density Estimation (KDE) curves of difference. While the second row shows the heatmap of the difference.} 
 	\label{fig:solutions}
\vspace{-0.5cm}
\end{figure}

\begin{figure*}[!tbb]
	\centering
	\includegraphics[width=1\textwidth]{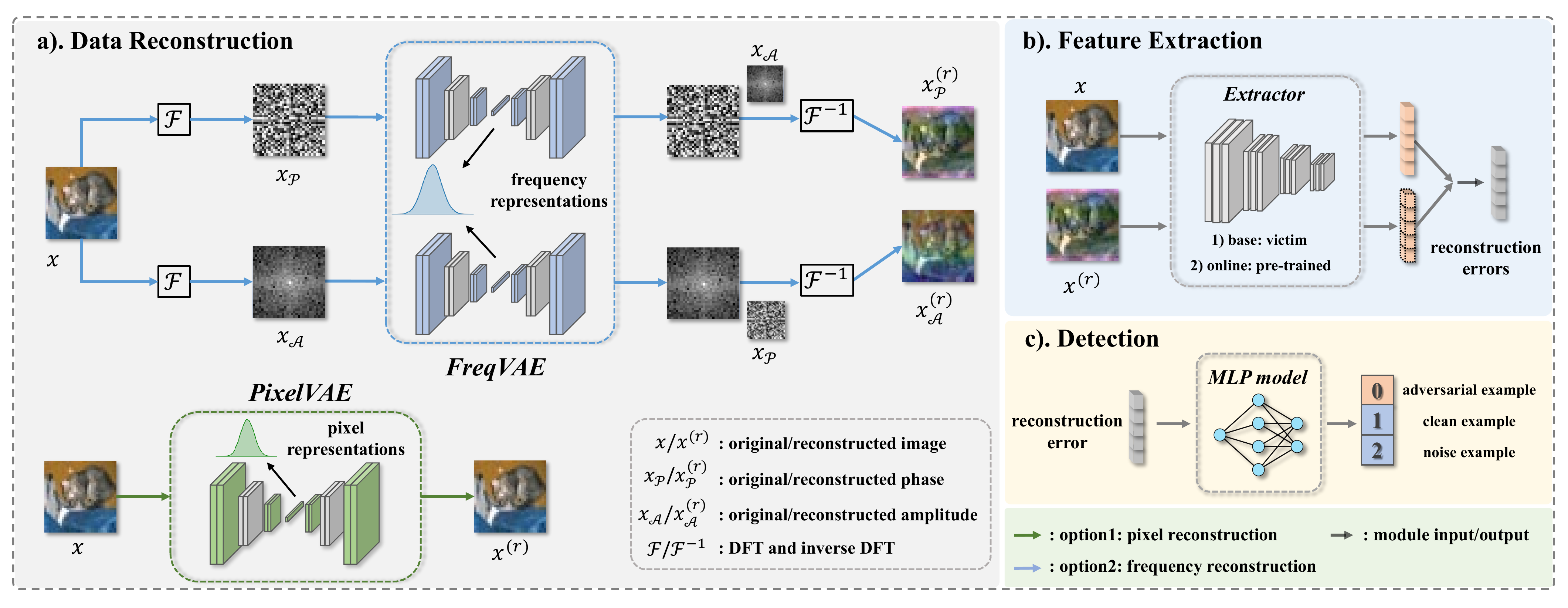}
    \caption{Overview of the proposed method consisting of (a) Data reconstruction, which reconstructs the original input images; (b) Feature extraction, which extracts the layer activations of the images before and after reconstruction; and (c) Adversarial example detection, which discriminates between the normal and adversarial examples through the reconstruction errors.} 
 	\label{fig:framework}
\end{figure*}

\noindent\textbf{Hypothesis behind performance degradation}: Based on the above observations, we hypothesize that the over-reliance on the victim model leads to the performance degradation. We observe that existing methods rely on the victim models for detector training in two ways:1) the victim model provides the training adversarial examples, which contain the victim model knowledge (e.g. the gradient information) and 2) the victim model provides the layer activations, which are directly used to design the detector features. Either way of reliance on the victim model may result in a failure to detect adversarial examples when the testing threat model is different from the victim model. 

Our BAD settings provide two black-box variables to demonstrate how the threat model is different from the victim model (i.e., the architecture and the strategy). We observe that strategy-based threat models exhibit more significant degradation than architecture-based threat models (e.g. in the MD method, SM(AT) is on average $19.92\%$ lower than SM(NT)). Moreover, the degradation is more pronounced in the EM setting (e.g. in the MD method, EM(AT) is on average $22.38\%$ lower than EM(NT)). Considering that adversarial training hardens the model knowledge (e.g., model robustness) and model ensemble increases the diversity of model knowledge, these factors directly result in more differences between the victim and threat model. This reveals that the performance degradation caused by model differences reflects the reliance on the victim model.

The above analyses demonstrate the importance of black-box adversarial examples for detector evaluation and attribute the performance degradation of existing detectors to the over-reliance on the victim model. This inspires us to develop effective BAD solutions by well circumventing the reliance on the victim models.

\subsection{Solution Inspiration}\label{Sec2.3}
It recent studies, data-based detection methods have been widely adopted to address the problem of out-of-distribution (O.O.D) detection \cite{zhou2022rethinking} and anomaly detection \cite{tian2019learning}, These methods do not require counterexamples or layer activations from the victim model during the detector training phase. Motivated by this, we aim to investigate the differences between normal and adversarial examples from a data perspective. The hypothesis behind is that the inherent differences in data patterns between normal and adversarial examples can be leveraged to tackle the BAD problem. To this end, we conducted preliminary experiments to analyze these differences in both the pixel and frequency domains.

\begin{equation}
    \label{difference}
    \left\{
    \begin{aligned}
        d_{o, i}^{\mathcal{D}} = \max\{\mathcal{M}(x)_{o} - \mathcal{M}(x\oplus\delta_{i}^{\mathcal{D}})_{o},\; \xi\} \\
        d_{t, i}^{\mathcal{D}} = \max\{\mathcal{M}(x\oplus\delta_{i}^{\mathcal{D}})_{t}-\mathcal{M}(x)_{t},\; \xi\}
    \end{aligned}
    \right.
\end{equation}

\noindent\textbf{Difference measurement}: The difference between normal and adversarial examples are defined as the change in prediction confidence caused by a perturbation. Specifically, we measure this impact by calculating the change in logits scores with respect to both the correct and target labels. Following \cite{xu2019interpreting}, the impact can be decomposed into two aspects: (1) the promotion of the target labels ($d_{o,i}$) and (ii) the suppression away from the correct labels ($d_{t,i}$). To compute $d_{o,i}$ and $d_{t,i}$ for an input image $x$ and its corresponding adversarial noise $\delta_{i}$, we first divide them into $m$ patches. We define $d_{o,i}$ and $d_{t,i}$ according to \eqref{difference}, where $\mathcal{M}(x)$ denotes the logits value, $o, t$ represent correct and target label, respectively, and $\xi >0$ is a constant value. In the pixel domain, we add the $i^{th}$ perturbation patch directly onto the input image. In the frequency domain, we replace the $i^{th}$ patch of a normal example with the phase and amplitude spectrums of the corresponding adversarial patch, respectively.

\vspace{0.15cm}

\noindent\textbf{Data Analysis}: We explore the differences between white- and black-box adversarial examples and project the difference value of each patch onto the original image in the form of a heatmap, where the activation regions correspond to the difference. As shown in the bottom panel of Figure \ref{fig:solutions}, we observe that the difference between normal and adversarial examples is evident in both the pixel and frequency domains, which confirms our hypothesis. An intriguing observation is that frequency domain reveals more activation regions than the pixel domain. To investigate whether this phenomenon is coincidental or not, we perform Kernel Density Estimation (KDE) on the difference values across the dataset. The results, shown in the top panel of Figure \ref{fig:solutions}, reveal that the mean value of the difference in the frequency domain (blue) is greater than in the pixel domain (red). This phenomenon is consistent across different black-box examples and may suggest that the frequency domain is a better choice for detecting adversarial example.

\section{Methodology}

\subsection{Overview and Preliminaries}

\noindent\textbf{Overview of data reconstruction-based solution}: Previous analyses have highlighted the potential of exploring BAD solutions from a data perspective. In this section, we introduce a data reconstruction-based adversarial example detection method. The pipeline of our method is illustrated in Figure \ref{fig:framework} and comprises three major modules: (1) Data reconstruction. We propose a simple yet effective VAE approach to learn versatile visual representations of normal examples, i.e., frequency (phase and amplitude spectrums, marked with blue color) or pixel-based representations (marked with green color, see Sec. \ref{Sec. 3.2}). (2) Feature extraction. This module provides the layer activations, which are subsequently used to calculate the reconstruction error (Sec. \ref{Sec. 3.3}). (3) Adversarial example detection. We employ the reconstruction error to identify the adversarial examples (Sec. \ref{Sec. 3.4}).

\vspace{0.15cm}

\noindent\textbf{Preliminaries}: Regarding frequency-based visual representation, two-dimensional images can be converted into frequency domain by performing the Fourier transform. We use 2D-Discrete Fourier Transform (DFT) and inverse DFT (IDFT) for the transformation between the pixel and frequency domains. Given an image $x\in R^{H\times W}$, where $H$ and $W$ denote the height and width of the image, let $\mathcal{R}(x)$ and $\mathcal{I}(x)$ be the real and imaginary part of $\mathcal{F}(x)$. The amplitude and phase components of a complex frequency value can be expressed as: $\mathcal{A}(x) = (\mathcal{R}(x)^2+\mathcal{I}(x)^2)^{1/2}$ , $\mathcal{P}(x)=arctan[\,\mathcal{I}(x) \,/\, \mathcal{R}(x)\,]$, respectively. Then we have $\mathcal{F}(x) = \mathcal{A}(x)\otimes e^{i\cdot \mathcal{P}(x)}$.

\begin{table*}[!tbp]

    \renewcommand\arraystretch{1.3}
    \centering
        \caption{Comparison of detection performance on CIFAR-10 and CIFAR-100 dataset under white- and black-box settings. \textbf{Bold} and \underline{underline} denote the best and second-best AUC scores under each setting. \textbf{\color{red} Red}: the increase compare to the best baseline method, SID}
        \vspace{0.15cm}
        \resizebox{0.9\textwidth}{!}{
            \setlength\tabcolsep{15pt}

            \scalebox{2}{
            \begin{tabular}{cccccccc}
                
                \toprule
                
                \makebox[0.1\textwidth][c]{\multirow{2}{*}{\Large Dataset} }
                &\makebox[0.18\textwidth][c]{\multirow{2}{*}{\Large Method} }
                &\makebox[0.08\textwidth][c]{\makebox[0.08\textwidth][c]{\multirow{2}{*}{White-box}}}
                &\multicolumn{4}{c}{\makebox[0.16\textwidth][c]{BAD Evaluation}} 
                &\makebox[0.08\textwidth][c]{\multirow{2}{*}{\Large Avg.}} \\ 
                    \cline{4-7}
                ~ & ~ & ~ & SM-AT & SM-NT & EM-AT & EM-NT & ~ \\
                    
                    \midrule[1.0pt]
                
                \multirow{11}{*}{\Large CIFAR-10}
                & LID \cite{DBLP:conf/iclr/Ma0WEWSSHB18} & 76.75 & 57.32 & 71.19 & 50.89 & 59.94 & 63.22 \\
                & MD \cite{lee2018simple}  & 87.67 & 60.40 & 80.32 & 48.39 & 70.77 & 69.51 \\
                & SID \cite{tian2021detecting} & 89.02 & 57.98 & 79.69 & 54.78 & 77.14 & 71.72 \\
                \cline{2-8}

                &    PRD-base & 91.76 & 67.25 & 88.94 & 51.75 & 85.13 & 77.22 (\textbf{\color{red}+5.50}) \\
                
                &   PRD-online & 87.08 & 71.52 & 88.58 & 66.51 & 87.56 & 80.25 (\textbf{\color{red}+8.53}) \\
                \cline{2-8}
                
                &    FRD-base (joint) & 91.93 & 67.27 & 89.89 & 53.03 & 85.65 & 77.55 (\textbf{\color{red}+5.83}) \\
                
                &    FRD-base (amp) & \underline{92.14} & 66.67 & 90.02 & 52.88 & 85.96 & 77.53  (\textbf{\color{red}+5.81}) \\ 
                
                &    FRD-base (pha) & 92.13 & 67.24 & 89.49 & 55.04 & 86.69 & 78.12  (\textbf{\color{red}+6.70}) \\
                \cline{2-8}
                
                &   FRD-online (joint) & 91.59 & \underline{73.43} & \underline{91.53} & \underline{66.77} & \underline{91.58} & \underline{82.98}  (\textbf{\color{red}+11.26}) \\
                
                &   FRD-online (amp) & \textbf{92.91} & 72.63 & \textbf{92.10} & 65.47 & \textbf{92.16} & \textbf{83.05}  (\textbf{\color{red}+11.33}) \\ 
                
                &   FRD-online (pha) & 90.09 & \textbf{73.73} & 90.52 & \textbf{66.82} & 89.59 & 82.15  (\textbf{\color{red}+10.43}) \\ 
                
                \hline
                \hline
                
                \multirow{11}{*}{\Large CIFAR-100}
                & LID \cite{DBLP:conf/iclr/Ma0WEWSSHB18} & 67.53 & 49.24 & 65.05 & 53.53 & 52.12 & 57.49 \\
                & MD \cite{lee2018simple}  & 82.32 & 59.02 & 74.97 & 61.09 & 63.04 & 68.09 \\
                & SID \cite{tian2021detecting}& 80.70 & 66.01 & 80.17 & 63.79 & 72.27 & 72.59 \\
                
                \cline{2-8}
                &    PRD-base & 90.91 & 63.41 & 89.99 & 75.85 & 76.76 & 79.38 (\textbf{\color{red}+6.79}) \\
                &   PRD-online & \textbf{94.82} & \underline{67.76} & \underline{93.07} & 74.54 & 77.91 & 81.62 (\textbf{\color{red}+9.03}) \\
                
                \cline{2-8}
                &    FRD-base (joint) & 92.10 & 66.48 & 91.09 & \underline{78.92} & \underline{80.91} & 81.90 (\textbf{\color{red}+9.31}) \\ 
                
                &    FRD-base (amp) & 92.16 & 66.33 & 91.16 & \textbf{79.75} & \textbf{81.67} & 82.22 (\textbf{\color{red}+9.63}) \\
                
                &    FRD-base (pha) & 91.09 & 64.65 & 89.66 & 78.20 & 80.04 & 80.73 (\textbf{\color{red}+8.14}) \\
                \cline{2-8}
                
                &   FRD-online (joint) & 93.64 & 67.72 & 91.78 & 77.35 & 80.00 & 82.10 (\textbf{\color{red}+9.51}) \\
                
                &   FRD-online (amp) & \underline{94.67} & \textbf{68.59} & \textbf{93.43} & 77.29 & 79.96 & \textbf{82.79} (\textbf{\color{red}+10.20}) \\
                
                &   FRD-online (pha) & 94.13 & 67.60 & 92.67 &  77.86 & 80.20 & \underline{82.49} (\textbf{\color{red}+9.90}) \\ 
                
                \bottomrule
                
                \end{tabular}
                }  
        }
        \vspace{-0.3cm}
\label{tab:comparisons}
\end{table*}

\subsection{Data Reconstruction}\label{Sec. 3.2}
\noindent\textbf{Pixel-based variational auto-encoder (PixelVAE)} has an encoder (parameterized by $\phi$) with multiple convolutional layers, which maps the original pixel signals $x$ to the latent representations $z$, and a decoder (parameterized by $\theta$), which reconstructs the original signals from the latent representations. Our loss function is defined as:

\begin{equation}
    \min\limits_{\phi, \theta}\,\, E_{q_{\phi}(z|x)} \log p_{\theta}(x|z) + \beta D_{\mathbb{KL}}(q_{\phi}(z|x) || p(z))
\end{equation}\label{VAE} 

\noindent where $p(z)$ is the prior of $z$ and $D_{\mathbb{KL}}(\cdot | \cdot)$ is the Kullback-Leibler divergence. This end-to-end training procedure can effectively encode features of normal examples into the latent space distribution, which is beneficial the downstream detection task. The rationale is that the PixelVAE module reconstructs normal examples with minimal reconstruction errors, indicating that the learned representations are effective in capturing the features of normal examples. However, when presented with potential adversarial examples, the PixelVAE module forces the corresponding latent representations shift towards the distribution of normal example, leading to significant increases in reconstruction errors. 

\vspace{0.15cm}

\noindent\textbf{Frequency-based variational auto-encoder (FreqVAE)} aims at reconstructing the original frequency signals of normal examples. Frequency signals contain the underlying patterns of images, which are critical for recognizing subtle but systematic modifications such as adversarial perturbations \cite{xie2022masked}. To achieve this, we perform DFT on normal examples to obtain the phase and amplitude spectrums. The amplitude spectrum mainly retains the low-level statistics, while the phase spectrum mainly captures the high-level structural information. The phase and amplitude spectrums are separately reconstructed by two VAE models. For example, in the amplitude spectrum reconstruction, we use the FreqVAE module to obtain the reconstructed amplitude spectrum, which is then combined with the original phase spectrum. We perform IDFT to obtain the amplitude-reconstructed image in the pixel domain. The loss function is similar to Eq. \eqref{VAE}, where $z$ corresponds to the latent representations of the original amplitude spectrum. It is noted that we compute the mean squared error (MSE) between the reconstructed and original images in the pixel domain, which is represented by the first term of Eq. \eqref{VAE}). Similarly, we keep the amplitude spectrum and reconstruct the phase spectrum in another FreqVAE module.

\subsection{Feature Extraction}\label{Sec. 3.3}
We adopt the approach outlined in Sec. \ref{Sec2.3} where the layer activations from the Convolutional Neural Networks (CNNs) model aer used as the detector features. It has been observed that model-driven adversarial perturbations result in image features being perceived by CNN models, allowing target class information can be easily expressed \cite{szegedy2013intriguing}. However, when the structure of adversarial perturbation is destroyed, the adversarial examples may fail to attack the CNN models \cite{jia2019comdefend}. Motivated by this phenomenon, we use a CNN model to extract the features of the images before and after reconstruction. We believe that this design does not contradict our original intention of eliminating the reliance on the victim model. Therefore, the architecture of feature extractor can be flexibly served by any CNN models.

\subsection{Adversarial Example Detection}\label{Sec. 3.4}
The proposed data reconstruction-based detection method (PRD and FRD) involves three main steps. Firstly, the input images are reconstructed using the PixelVAE or FreqVAE module. Secondly, the layer activations of the images before and after reconstruction are captured through a feature extraction module. The victim model is used for feature extraction, resulting in the base version of the method. To further alleviate the reliance on the victim model, a pre-trained model (not the victim model) is used instead, resulting in the online version of the method. The online version provides a unified protection for victim models with different structures, making it easier to deploy real-world models. Finally, the reconstruction error is calculated using the layer activation, defined as: $\Vert \mathcal{M}_{l}(x_{r}) - \mathcal{M}_{l}(x_{o}) \Vert$, where $x_{r}$ and $x_{o}$ corresponds to the image before and after reconstruction. $\mathcal{M}_{l}$ is the $l^{th}$ layer activation. The reconstruction error is then passed into a detector, such as a MLP model.

\vspace{0.15cm}

\noindent\textbf{Training details}: The training algorithm of proposed methods is available in the appendix. We obtain two variants of methods for PRD and FRD, shorten as PRD-base, PRD-online, FRD-base and FRD-online. For each FRD method, we consider the following three types of reconstructed images: (a) phase representations (pha): $\mathcal{F}^{-1}(\mathcal{A}_{o} e^{i\cdot \mathcal{P}_{r}})$; (b) amplitude representations (amp): $\mathcal{F}^{-1}(\mathcal{A}_{r}  e^{i\cdot \mathcal{P}_{o}})$; and (c) joint representations (joint): $\mathcal{F}^{-1}(\mathcal{A}_{r} e^{i\cdot \mathcal{P}_{r}})$, where $\mathcal{A}_{o}, \mathcal{P}_{0}, \mathcal{A}_{r}$, and $\mathcal{P}_{r}$ denote the amplitude and phase spectrums before and after reconstruction. Thus we obtain three variants of each FRD method, shorten as FRD (pha), FRD (amp) and FRD (joint). Similar to the prior researches, normal examples include clean and noisy examples, both of them are correctly classified by the victim model. The noisy examples are produced by adding random noises with the same magnitude of adversarial perturbations. In detector training phase, we prepare a detection dataset consisting of examples from the above three classes with labels 0, 1 and 2. In detector testing phase, class 1 and class 2 are merged.

\begin{table}[!tp]
    \renewcommand\arraystretch{1.3}
    \centering
        \caption{Detection performance comparison among different feature extractors under BAD evaluation. Here we consider three pre-trained models: ResNet34, MAE\cite{he2022masked} and SimCLR\cite{pmlr-v119-chen20j}}
        \vspace{0.15cm}
        \resizebox{0.48\textwidth}{!}{%
        \setlength\tabcolsep{7pt}
        \scalebox{0.5}{
        \begin{tabular}{ccccccc}
            
            \toprule
            
            \makebox[0.05\textwidth][c]{\multirow{2}{*}{\Large Method } }
            &\makebox[0.08\textwidth][c]{\makebox[0.08\textwidth][c]{\multirow{2}{*}{White-box}}}
            &\multicolumn{4}{c}{\makebox[0.16\textwidth][c]{BAD Evaluation}}  
            &\makebox[0.03\textwidth][c]{\multirow{2}{*}{\Large Avg.}} \\ 
                \cline{3-6}
            ~ & ~ & SM-AT & SM-NT & EM-AT & EM-NT & ~ \\
                
            \midrule[1.0pt]
            \textbf{Baseline} & 89.02 & 60.40  & 80.32 & 54.78 & 77.14 & 72.33 \\
            MAE & 73.65 & 70.49 & 75.69 & 83.87 & 81.27 & 76.99 (\textbf{\color{red}+4.66})\\
            ResNet34 & 97.70 & 67.12 & 95.29 & 67.26 & 94.20 & 84.31 (\textbf{\color{red}+11.98})\\
            SimCLR & 95.97 & 70.89 & 92.62 & 90.62 & 75.20 & 85.06  (\textbf{\color{red}+12.73})\\           
            \bottomrule
            \end{tabular}
    }
}
\vspace{-0.35cm}
\label{tab:different-extractors-selection}
\end{table}

\section{Experiment Results}
\subsection{Experiment Setup}

\noindent\textbf{Datasets and models}: We experiment on CIFAR-10 and CIFAR-100 datasets. Settings of the victim and threat models are shown in Table \ref{tab1:black-setting}. For PRD and FRD methods, we set ResNet18 (victim model) and WideResNet28 (WRN28, pre-trained model) as the feature extraction module for each base and online variant, respectively. We consider two fully-connected layers as the structure of the detector. The structures of PixelVAE and FreqVAE, and other training details are available in the appendix.

\vspace{0.15cm}

\noindent\textbf{Baseline models}: Performance of PRD and FRD are compared with the state-of-the-art adversarial example detection methods based on local intrinsic dimensionality (LID), Mahalanobis detector (MD) and sensitivity inconsistency detector (SID). Each detection method is implemented with six detectors trained on FGSM, BIM, PGD-$l_{2}$, PGD-$l_{\infty}$, DeepFool and C\&W examples (provided by the victim model). The final detection performance is obtained by averaging the results of the above six detectors.

\vspace{0.15cm}

\noindent\textbf{Evaluation metrics}: The metric for evaluation is the widely used AUC score. For fair comparison, all these methods share the same evaluation environment and all the parameters are fine-tuned to achieve the best performance.

\subsection{Performance Comparison Under BAD}
Table \ref{tab:comparisons} summarizes the performance of all detectors on CIFAR-10 and CIFAR-100 datasets. On both datasets, our proposed methods (PRD and FRD) outperform LID, MD and SID in both black- and white-box settings, providing effective solutions to address BAD. It is noted that the overall increase in PRD-base variant is less significant than FRD-base variants, e.g., the average detection performance of FRD-base variants (i.e., average among the joint, amp and pha variants) is increased by $0.61\%$ and $2.24\%$ on CIFAR-10 and CIFAR-100 datasets. This indicates that the FreqVAE module is better at capturing the separability between normal and adversarial examples, which is also consistent with our observations in Sec. \ref{Sec2.2}. Moreover, we have the following intriguing and valuable observations: 

\vspace{0.15cm}

\noindent\textbf{On frequency representations}: We conduct a comparison among the three FRD variants: FRD (joint), FRD (pha) and FRD (amp). Here we both consider base and online variants. We observe that (1) performance of the three detectors are very close, and (2) the best performance is achieved by different frequency representations in different black-box settings. Considering that the FreqVAE module for amplitude and phase spectrum are trained separately, we show that both of them can explore the separability between normal and adversarial examples independently, and play critical roles in solving different black-box settings.

\vspace{0.15cm}

\noindent\textbf{On feature extractor}: When applying a pre-trained model as feature extraction module, PRD- and FRD-online methods outperform the base versions remarkably on both white- and black-box settings. For example, in CIFAR-10 detection, the average performance of PRD-online is $3.03\%$ higher than PRD-base, and FRD-online is $4.89\%$ higher than FRD-base. This suggests that the online version can effectively relieve the need for the victim model (regarding the layer activations, see Sec. \ref{Sec2.2}). Therefore, we shed light on the advantage of our online version in real-world deployments, which lies in providing unified protection for the victim models with flexible structures. To further demonstrate this advantage, Table \ref{tab:different-extractors-selection} reports the detection performance of the other three pre-trained models as feature extraction modules. We observe that the average detection performances of all models outperform than the baseline model. The detailed settings are available in the appendix.

\begin{table}[!tbp]
    \renewcommand\arraystretch{1.3}
    \centering
        \caption{Comparison in real-world black-box attack detection. Each method considers the best detectors in BAD evaluation.}
        \vspace{0.25cm}
        \resizebox{0.35\textwidth}{!}{%
        \setlength\tabcolsep{13pt}
        \scalebox{0.3}{
        \begin{tabular}{cccccc}
            \toprule
            Method & CGATTACK\cite{feng2022boosting} & TAIG\cite{huang2022transferable} \\
            \midrule
            LID & 85.13 & 61.69 \\
            MD & 95.59 & 74.52 \\
            SID & 84.44 & 90.46 \\
               PRD (ours) & 97.90 & 91.56 \\
              FRD (ours) & \textbf{98.56} & \textbf{93.81} \\
            
            \bottomrule
            \end{tabular}
    }
}
\vspace{-0.35cm}
\label{tab:with-real-world-black-box-attack}
\end{table}

\subsection{Real World Black-box Attack Detection}
Previous subsection has shown an excellent detection performance of PRD and FRD under BAD evaluation. To further demonstrate the capabilities of PRD and FRD in detecting black-box attacks, we evaluate the real-world black-box attack detection on CIFAR-10 dataset, using ResNet-18 as the victim model. Here we consider two advanced black-box attacks, CGATTACK \cite{feng2022boosting} and TIAG \cite{huang2022transferable}. Table \ref{tab:with-real-world-black-box-attack} summarizes the detection performance for different methods. We observe that both PRD and FRD outperform LID, MD and SID in detecting the above two black-box attacks detection, which is consistent with the BAD evaluation and confirms the effectiveness of our method. Although baseline methods have good detection results in some black-box attacks, this is not always true, e.x., MD does well in the CGATTACK detection but almost fails in the TAIG detection. This further proves the limitation in the existing detection methods: the reliance on the victim model.

\begin{table}[!tbp]
    \renewcommand\arraystretch{1.3}
    \centering
        \caption{Performance comparison with CD-VAE detection method under BAD evaluation.}
        \vspace{0.15cm}
        \resizebox{0.48\textwidth}{!}{%
        \setlength\tabcolsep{7pt}
        \scalebox{0.5}{
        \begin{tabular}{cccccc}
            \toprule
            
            \makebox[0.05\textwidth][c]{\multirow{2}{*}{\Large Method} }
            &\makebox[0.08\textwidth][c]{\makebox[0.08\textwidth][c]{\multirow{2}{*}{White-box}}}
            &\multicolumn{4}{c}{\makebox[0.16\textwidth][c]{BAD Evaluation}} \\ 
                \cline{3-6}
            ~ & ~ & SM-AT & SM-NT & EM-AT & EM-NT \\
            
            \midrule
            CD-VAE \cite{NEURIPS2021_CDVAE} & 95.23 & 74.16 & 92.97 & 58.43 & 82.99 \\
            FreqVAE & \textbf{98.38} & \textbf{79.02} & \textbf{96.46} & \textbf{78.21} & \textbf{98.53} \\
            \bottomrule
            \end{tabular}
    }
}
\label{tab:with-CD-VAE}
\end{table}

\vspace{0.25cm}

\begin{table}[!tbp]
    \renewcommand\arraystretch{1.3}
    \centering
        \caption{Detection performance FRD-base (pha) variant on different perturbation strengths.}
        \vspace{0.15cm}
        \resizebox{0.48\textwidth}{!}{%
        \setlength\tabcolsep{7pt}
        \scalebox{0.5}{
        \begin{tabular}{ccccccc}
            
            \toprule
            
            \makebox[0.05\textwidth][c]{\multirow{2}{*}{\Large Strength} }
            &\makebox[0.08\textwidth][c]{\makebox[0.08\textwidth][c]{\multirow{2}{*}{White-box}}}
            &\multicolumn{4}{c}{\makebox[0.16\textwidth][c]{BAD Evaluation}} 
            &\makebox[0.03\textwidth][c]{\multirow{2}{*}{\Large Avg.}} \\ 
                \cline{3-6}
            ~ & ~ & SM-AT & SM-NT & EM-AT & EM-NT & ~ \\
                
            \midrule[1.0pt]
            2/255 & 81.83 & 75.06 & 59.97 & 87.23 & 82.61 & 77.34 \\
            4/255 & 87.19 & 74.91 & 76.25 & 85.67 & 94.73 & 83.75 \\
            6/255 & 92.23 & 75.54 & 85.30 & 77.03 & 95.80 & 85.18 \\
            8/255 & 95.41 & 70.66 & 91.99 & 57.41 & 92.58 & 81.61 \\
            
               Total & 85.23 & 77.25 & 73.01 & 86.92 & 93.02 & 83.09 \\
              CTR-balance & 90.06 & 76.14 & 83.02 & 85.99 & 95.76 & \textbf{86.19} \\
            
            \bottomrule
            \end{tabular}

    }
}
\vspace{-0.15cm}
\label{tab:perturbation-strength}
\end{table}

\subsection{Comparison with Other Reconstruction Based Detection Methods}
Here we elaborate on the connections to the most related reconstruction based detection works CD-VAE \cite{NEURIPS2021_CDVAE} and MagNet \cite{meng2017magnet}.

\noindent\textbf{Relation to MagNet}: MagNet is an early work to detect the adversarial examples by reconstruction error. However, MagNet significantly diverges from our work in two fundamental aspects: (1) The assumption on black-box definition is based on the attacker’s full access to the victim model, which violates the assumption of our BAD definition. (2) MagNet is an unsupervised detection approach, where PRD and FRD are supervised detection method.

\vspace{0.15cm}

\noindent\textbf{Relation to CD-VAE}: CD-VAE is another related work on data reconstruction that proposes a detection-defense framework. However, CD-VAE cannot be served as a plugin, thus fails to provide detection-based protection for the deployed models in real-world scenarios. Despite this limitation, we still evaluate the detection performance improvement brought by the reconstruction module. We replace the FreqVAE module with CD-VAE and conduct the same experiments without changing the feature extraction and detection modules. Table \ref{tab:with-CD-VAE} indicates that our method still outperforms CD-VAE in the BAD problem, confirming its effectiveness.

\begin{figure}[!tbp]
\centering\includegraphics[width=0.48\textwidth]{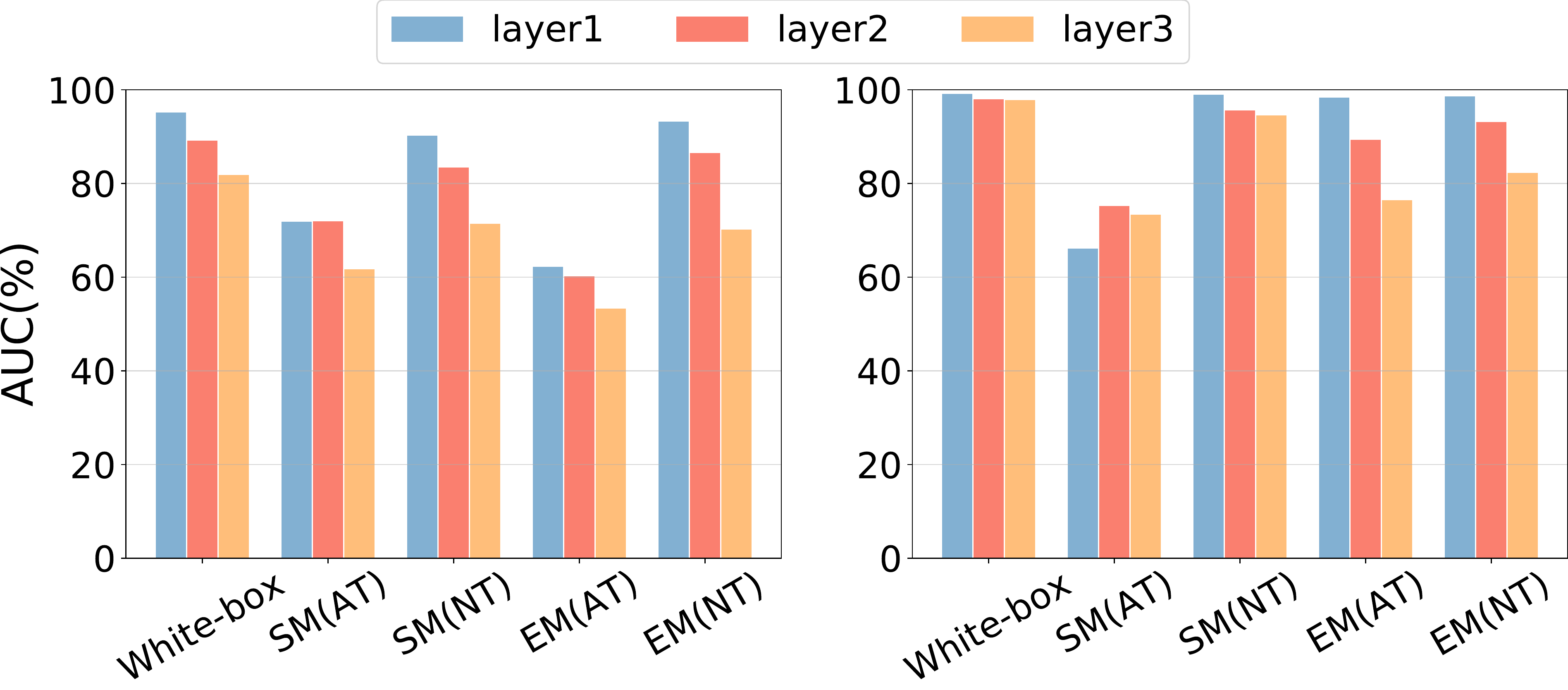}
    \caption{Detection performance of FRD method on different layer activations. Here we consider the best two detectors: (left) PGD-$l_{\infty}$ detector in FRD-base (pha) variant and (right) PGD-$l_{2}$ detector in FRD-online (amp) variant.} 
    \label{fig:section5}
\end{figure}

\subsection{Parameter Influence}
To more thoroughly show the performance details of our proposed methods, in this subsection, we report the analysis results of parameter influence from the perspective of layer activations and perturbation strengths. Experiments are carried out based on the FRD method.

\vspace{0.15cm}

\noindent\textbf{On layer activations}: We analyze the effect of activation from different layers of the feature extraction module on the detection performance. We compare the best two detectors: PGD-$l_{\infty}$ detector (FRD-base (pha) variant, Figure \ref{fig:section5}.left) and PGD-$l_{2}$ detector (FRD-online (amp) variant, Figure \ref{fig:section5}.right). The key observation is that features extracted from the top and middle layers result in a better detection performance compared to the bottom layers in most cases. This suggests that the reconstruction errors of low-level features (e.g., edge, color) are more sensitive to adversarial perturbations than high-level features (e.g., abstract semantic information). This is in contrast with LID, MD and SID, whose typical selection is usually at the bottom layers.

\vspace{0.15cm}

\noindent\textbf{On perturbation strengths}
To select a suitable perturbation for detector training, we set some pre-defined strengths ranging from 2/255 to 8/255. We report the AUC scores of PGD-$l_{\infty}$ detector (FRD-base (pha) variant) trained on these strengths. The detection results on CIFAR-10 dataset are shown in Table \ref{tab:perturbation-strength}. We see that in the case of strategy-based attacks (e.g., EM(AT) setting), the AUC scores rapidly decrease when the perturbation strength becomes larger. However, this tendency is not observed in the architecture-based attack (e.g., SM(NT) setting), where a large perturbation strength achieves superior performance. The perturbation strength provides the trade-off among different black-box settings, suggesting that one perturbation training strategy is not sufficient for addressing BAD.

\subsection{Analysis of Data Reconstructed Images}
In order to figure out what factors lead to the separability between normal and adversarial examples, we compare and analyze the images before and after reconstruction. Experiment will be carried out based on the FreqVAE module.

\vspace{0.15cm}

\noindent\textbf{On feature similarity}: We extract the features (layer activations) of the images before and after reconstruction and calculate the cosine similarity as shown in Table \ref{tab:cosine-similarity}. For normal examples, the average similarity is approximately 0.9, indicating the capability of the FreqVAE module in capturing the frequency representations of normal examples. However, the average similarity of adversarial examples is much lower for both white- and black-box settings, suggesting that the FreqVAE module modifies adversarial examples. Such phenomena reveals the capability difference of the FreqVAE module in image reconstruction, motivating us to further explore how the capability difference affects the above separability.

\vspace{0.15cm}

\begin{figure}[!tbb]
	\centering

	\includegraphics[width=0.47\textwidth]{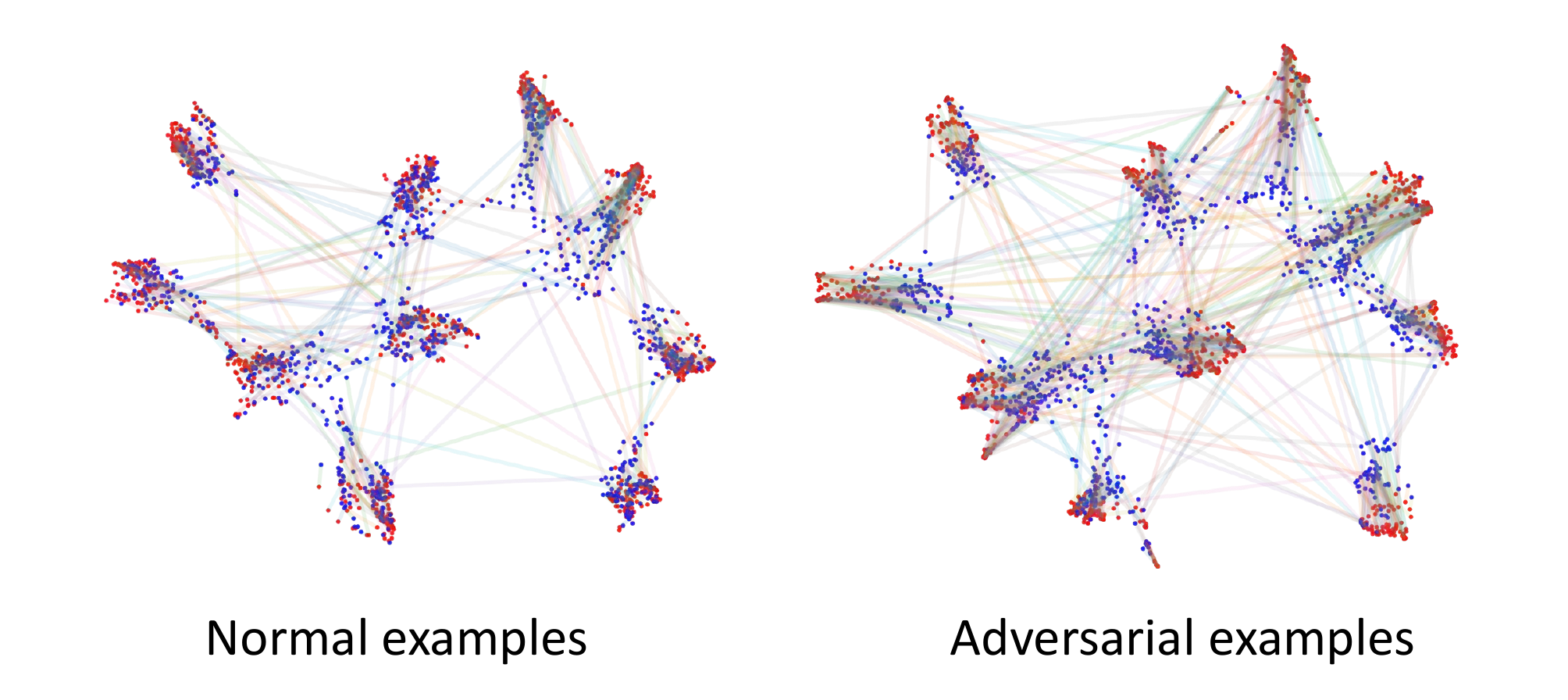}
    \caption{Feature visualization through umap} 
 	\label{fig:umap}
\end{figure}

\begin{table}[!t]
    \renewcommand\arraystretch{1.3}
    \centering
        \caption{Similarity of images before and after reconstruction.}
        \vspace{0.25cm}
        \resizebox{0.4\textwidth}{!}{%
        \setlength\tabcolsep{9pt}
        \scalebox{0.5}{
        \begin{tabular}{ccccc}
            
            \toprule
            White-box & SM-AT & SM-NT & SM-AT & SM-NT \\                
            \midrule     
            0.9012 & 0.8238 & 0.8560 & 0.8366 & 0.8476 \\
            \bottomrule
            \end{tabular}
    }
}
\label{tab:cosine-similarity}
\end{table}

\noindent\textbf{On class transfer rate (CTR)}: In Figure \ref{fig:umap}, we visualize these features through umap \cite{umap}. We find that normal examples (left) mainly exhibit inner-cluster transfer, while adversarial examples (right) exhibit inter-cluster transfer (as indicated by line density), which supports our observation on feature similarity. To further illustrate the cluster transfer, we introduce the predicted label information of the victim model. We observe two cases of class transfer: (1) label consistency (LC), where the predicted label is same before and after reconstruction, and (2) label inconsistency (LI) ,where the predicted label changes. Therefore, we define the CTR scores of LC and LI as LC/(LC+LI) and LI/(LC+LI), respectively. We analyze the CTR score of the adversarial examples with different perturbation strengths, as shown in Table \ref{tab:CTR}. For normal examples, the CTR score remains constant under different strengths and achieves an LC score of $80.23\%$, indicating that the reconstructed normal examples are still correctly classified. Moreover, we observe that in the case of small perturbation, the CTR score of LC is higher than that of LI, which suggests a feature shift towards the non-adversarial class distribution. We argue that the capability of the FreqVAE module reflects on two aspects: (1) maintaining the original normal class information and 2) suppressing the original adversarial class information in the case of the adversarial examples. The different ways in which the FreqVAE module handles the class information for normal and adversarial examples directly leads to the separability. However, we observe that the suppression effect on the adversarial class information is hindered under large perturbations. Nevertheless, we demonstrate that the conclusion still holds (Detailed in the appendix).

\vspace{0.15cm}

\begin{table}[!tbp]
    \renewcommand\arraystretch{1.3}
    \centering
        \caption{Class transfer rate of normal and adversarial examples.}
        \vspace{0.15cm}
        \resizebox{0.48\textwidth}{!}{%
        \setlength\tabcolsep{11pt}
        \scalebox{0.3}{
        \begin{tabular}{cccccc}
            
            \toprule

            \makebox[0.02\textwidth][c]{\makebox[0.08\textwidth][c]{\multirow{2}{*}{}}}
            &\makebox[0.08\textwidth][c]{\makebox[0.08\textwidth][c]{\multirow{2}{*}{normal}}}
            &\multicolumn{4}{c}{\makebox[0.08\textwidth][c]{adversarial}} \\
            \cline{3-6}
            
            ~ & ~ & 2/255 & 4/255 & 6/255 & 8/255 \\
            \midrule
            
            LC & 80.23 & 36.09 & 53.01 & 56.02 & 60.51 \\
            LI & 19.77 & 63.91 & 46.99 & 43.98 & 39.49 \\
            
            \bottomrule
            \end{tabular}
    }
}

\label{tab:CTR}
\end{table}

\noindent\textbf{Solution improvement}: For an ideal reconstruction module, any example should satisfy the class transferring rules (as discussed in previous subsection, i.e., normal examples exhibit inner-class transfer while adversarial examples exhibit inter-class transfer), which helps to explore the separability between normal and adversarial examples in the down-stream detection task. However, based on previous analyses, we find a decreasing tendency of inter-class transfer rate when the perturbation strength becomes larger. This reflects a potential limitation of the FreqVAE module, e.g., a detector trained on large perturbations (CTR score of LC is $60.51\%$ for 8/255) may have a performance degradation on those adversarial examples with small perturbations (CTR score of LI is $63.91\%$ for 2/255). 

To solve this challenge, we construct a new training dataset called CTR-balance. The improvement consists of two aspects: (1) We sample the adversarial examples under different perturbation strengths; and (2) We perform balanced sampling according to the above class-transfer cases (LC and LI). To avoid data scale being a confounding factor (i.e., better performance with more data), we construct another dataset called total which contains all examples under the four perturbation strengths. Experiment results are shown in Table \ref{tab:perturbation-strength}. The average performance of the CTR-balance dataset is the best. At the same time, we notice that data scale does not improve the detection performance. Within the limitation of FreqVAE reconstruction, we provide a CTR-balance sampling strategy, which effectively addresses our concerns and offers empirical guidance for tackling BAD.

\section{Conclusions}
In this paper, we have positioned the problem of black-box adversarial example detection (BAD) and observed a significant performance degradation of existing detection methods in BAD. Our observations and analyses have attributed this degradation to an over-reliance on the victim model, which has inspired us to design a data reconstruction-based adversarial example detection method in both pixel and frequency domains. Extensive experiments have demonstrated the effectiveness of our proposed method. There are still many problems that need to be addressed towards a practical BAD solution. Some of them are: (1) conducting more in-depth analysis regarding the reliance on the victim model, which could facilitate a better understanding of BAD and lead to the development of better BAD solutions; and (2) developing more comprehensive settings for BAD evaluation, such as considering additional attack-defense combinations as in the robust defense studies \cite{aldahdooh2022adversarial}.

{\small
\bibliographystyle{ieee_fullname}
\bibliography{main}
}

\appendix

\begin{figure*}[tbb]
	\centering
	\includegraphics[width=1\textwidth]{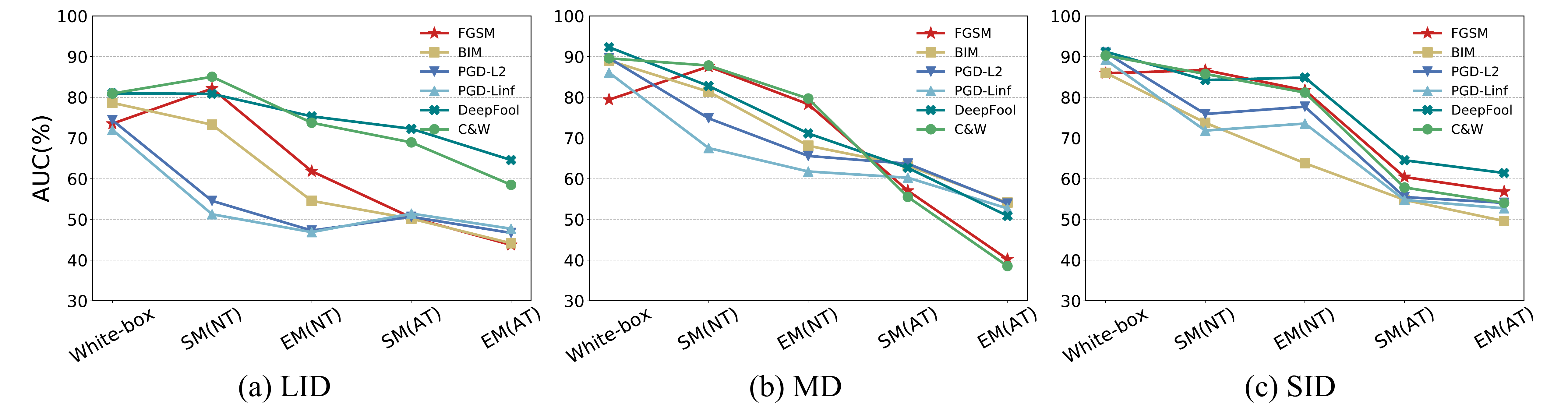}
    \vspace{-0.8cm}
    \caption{BAD evaluation on CIFAR-10 dataset.} 
 	\label{fig:data-cifar10}
\end{figure*}

\begin{figure*}[tbb]
	\centering
	\includegraphics[width=1\textwidth]{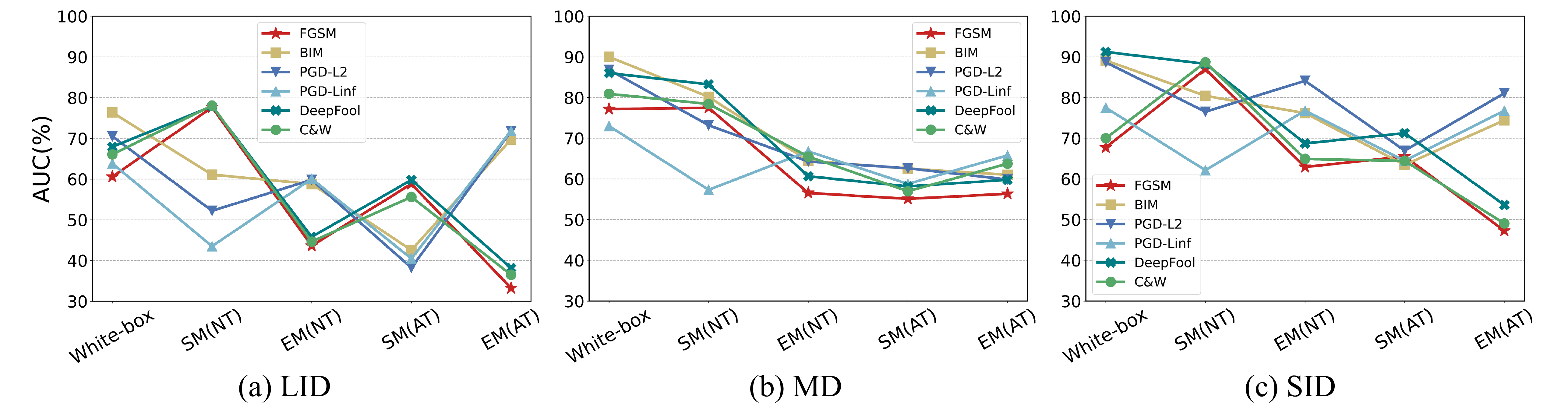}
    \vspace{-0.8cm}
    \caption{BAD evaluation on CIFAR-100 dataset.} 
 	\label{fig:data-cifar100}
\end{figure*}

\begin{table}[b]
    \huge
    \renewcommand\arraystretch{1.3}
    \centering
        \huge
        \caption{Binary accuracy of images before and after reconstruction.}
        \resizebox{0.3\textwidth}{!}{
        \setlength\tabcolsep{12pt}
        \scalebox{1}{
        \begin{tabular}{cc}
            \toprule
            normal example & adversarial example \\
            \midrule
            $78.36\%$ & $99.39\%$ \\
           \bottomrule
            
        \end{tabular}
        }
    }
\label{tab: binary-acc}
\end{table}

\section{BAD Evaluation}
Figure \ref{fig:data-cifar10} and \ref{fig:data-cifar100} show the detection performance (y-axis, AUC scores) evaluated in various black-box settings (x-axis). We also compared the performance in white-box setting. The key observations is that all detectors (line color) manifest performance degradation in black-box settings, which indicates that the existing detection methods cannot well transfer to BAD.

\section{Algorithm}
The training details of our proposed data reconstruction-based adversarial example detection method is given in Algorithm ~\ref{alg:algorithm}. For the training dataset $X_{train}$ of CIFAR-10 dataset, we capture the pixel or frequency representations through a VAE model. Given a clean image batch $B_{c}$ in the testing dataset $X_{val}$, we generate the adversarial image batch $B_{a}$. Similar to prior researches, the normal examples includes clean examples and noisy examples which are all correctly classified by the victim model. The noisy image batch $B_{n}$ can be readily produced by adding random noises with the same magnitude of adversarial perturbation. In detector training phase, we capture the layer activations from the $l^{th}$ layer of the feature extractor module. The reconstruction error are defined as $\vert \mathcal{M}_{l}(x_{r}) - \mathcal{M}_{l}(x_{o}) \vert$. Then the reconstruction error will be passed into a detector. Here we consider two fully-connected layer as the final detector.

\begin{algorithm}[tb]
    \caption{Training procedure of our proposed method}
    \textbf{Input}: Train data:$X_{train}$ and validation data:$X_{val}$ \\
    \textbf{Input}: $\mathcal{M}\triangleq$ Feature extraction module with $L$ layers \\
    \textbf{Output}: $F\triangleq$ Pixel/FreqVAE module, $\mathcal{D}\triangleq$ Detector
    
    \begin{algorithmic}[1] 
        \STATE $F\triangleq$ train Pixel/FreqVAE module with $X_{train}$
        \FOR{$B_{c}$ in $X_{val}$}
            \STATE $B_{a}\triangleq$ attack $B_{c}$ with attack method;
            \STATE $B_{n}\triangleq$ add random noise to $B_{c}$;
            \STATE $\rhd$ same perturbation magnitudes with $B_{a}$
            
            \STATE $\{R_{a};R_{c};R_{n}\}\triangleq F\{B_{a};B_{c};B_{n}\}$;
            
            \STATE $\rhd$ image reconstruction through $F$

            \FOR{$l$ in $[1, L]$}
                \STATE $rd_{l}^{+}\triangleq \{\vert\mathcal{M}_{l}(R_{a})-\mathcal{M}_{l}(B_{a})\vert \}$; 
                \STATE $rd_{l}^{-}\triangleq \{\vert\mathcal{M}_{l}(R_{x})-\mathcal{M}_{l}(B_{x})\vert, \;\;\, x \in \{c, n\}\}$; 
            \STATE $\rhd$ get reconstruction error in activation layer $l$
            \ENDFOR
            \STATE $D^{+}$.append($rd_{l}^{+}$), $D^{-}$.append($rd_{l}^{-}$)
        
        \ENDFOR
        \STATE $\mathcal{D}\triangleq$ train a binary classifier on $\{ D^{+}, D^{-} \}$
        \STATE $\rhd$ usually logistic regression (LR) model or MLP.
    \end{algorithmic}
    \label{alg:algorithm}
\end{algorithm}
\section{Experiment Setup}
\subsection{Black-box and White-box Setting}

\noindent\textbf{Victim and threat model settings}: For the problem of adversarial examples detection, we adopt six different models for classifying CIFAR-10 and CIFAR-100 datasets. We set ResNet18(NT) as the default victim model. For single-model (SM) attack, we set VGG16(NT), WResNet28(NT) and ResNet18(AT) as the threat model. For ensemble-model (EM) attack, we set VGG16(NT) \& WResNet28(NT) and VGG16(AT)\& WResNet28(AT) as the threat model, where \& means model ensemble.

\vspace{0.15cm}

\noindent\textbf{Test attacks settings}: We evaluate the detection performance on both white-box and black-box setting: (i) For white-box and SM setting, we consider six adversarial attacks: FGSM, BIM, PGD-$l_{2}$, PGD-$l_{\infty}$, DeepFool and C\&W. (ii) For EM setting, we consider only PGD-$l_{\infty}$ attack. Both two settings are performed non-targeted attack to get the adversarial examples.

\vspace{0.15cm}

\subsection{Adversarial Example Detector Setting}
\noindent\textbf{Pixel \& FreqVAE settings}: During the training process, we use SGD with a momentum of 0.9 and weight decay of 1e-4 as the optimizer to minimize Equation 2 in an end-to-end manner for 50 epochs on CIFAR-10 training set. We use three VAE models with a few convolutional layers to capture the phase, amplitude and pixel information of the normal example, respectively. It is noted that the FreqVAE trained on CIFAR-10 is suitable for both CIFAR-10 and CIFAR-100 detection task.

\vspace{0.15cm}

\noindent\textbf{PRD and FRD settings}: For PRD and FRD methods, we set ResNet18 (victim model) and WResNet28 (pre-trained model) as the feature extractor module for the base and online variants, respectively. We consider two fully-connected layers as the structure of the detector. On CIFAR-10 and CIFAR-100 dataset, the batch size is 128 and 256, respectively. Other default settings is available in Table \ref{tab: Setting-FRD},\ref{tab: Setting-PRD}.

\vspace{0.15cm}

\noindent\textbf{Various feature extractor selections setting}: In addition to the WideResNet28 model (used as feature extraction module in this paper), we also consider three additional pre-trained models: ResNet34, SimCLR, where we use ResNet18 as backbone and MAE, where we use ViT-tiny as backbone, as a supplement to prove the effectiveness of the online version. All these methods are trained on CIFAR-10 dataset. The baseline model is defined as the best result in different settings among LID, MD and SID.

\section{Experiment Results}
\subsection{Performance under BAD Evaluation}
We report the detailed performance of each detector on Table \ref{tab:comparisons_cifar10}, \ref{tab:comparisons_cifar100}. On both CIFAR-10 and CIFAR-100 datasets, our proposed methods (PRD and FRD) outperform LID, MD and SID on black-box as well as white-box settings, which provide effective solutions to address BAD. 

\subsection{Analysis of Reconstructed Images}
We observe that under large perturbations, the features (corresponds to the images before and after data reconstruction) exhibit far away from (but not out of) the original adversarial class distribution. We assume that although the reconstruction errors mainly correspond to the inner-class transfer in the case of large perturbations (e.x. 8/255 perturbation, the CTR score of LC is $60.51\%$). It is still distinguishable from normal examples (The CTR score of LC is $80.23\%$). To verify our hypothesis, we use a linear regression (LR) model to distinguish those inner-class transfer examples before and after reconstruction (including the normal and adversarial examples). The binary accuracy can reflect the difference in inner class transfer. As shown in Table\ref{tab: binary-acc}, the feature distinguishability of adversarial examples before and after reconstruction is significantly better than that of normal examples. That is, in the case of large perturbations, the reconstructed errors dominated by inner-class transfer can still guide the detection tasks.

\newpage

\begin{table*}[t]
    \huge
    \renewcommand\arraystretch{1.3}
    \centering
        \huge
        \caption{FRD setting for CIFAR-10 and CIFAR-100 detection.}
        \resizebox{0.9\textwidth}{!}{
        \setlength\tabcolsep{15pt}
        \scalebox{1}{
        \begin{tabular}{ccccccccccc}
            \toprule
             ~ & \multicolumn{5}{c}{CIFAR-10} & \multicolumn{5}{c}{CIFAR-100} \\ 
                \cline{2-11}
            ~ & \textbf{PGD} & \textbf{BIM} & \textbf{FGSM} & \textbf{DeepFool} & \textbf{C\&W} & \textbf{PGD} & \textbf{BIM} & \textbf{FGSM} & \textbf{DeepFool} & \textbf{C\&W} \\
           \midrule
            learning rate & 1e-2 & 1e-2 & 1e-2 & 1e-2 & 1e-2 & 5e-2 & 1e-2 & 1e-2 & 5e-2 & 1e-2 \\ 
            momentum  & 0.9 & 0.9 & 0.9 & 0.9 & 0.9 & 0.9 & 0.9 & 0.9 & 0.9 & 0.9 \\
            weight decay & 5e-4 & 5e-3 & 5e-3 & 5e-3 & 5e-3 & 5e-4 & 5e-3 & 5e-3 & 5e-4 & 5e-3 \\
           \bottomrule
            
        \end{tabular}
        }
    }
\label{tab: Setting-FRD}
\end{table*}

\begin{table*}[t]
    \huge
    \renewcommand\arraystretch{1.3}
    \centering
        \huge
        \caption{PRD setting for CIFAR-10 and CIFAR-100 detection.}
        \resizebox{0.9\textwidth}{!}{
        \setlength\tabcolsep{15pt}
        \scalebox{1}{
        \begin{tabular}{ccccccccccc}
            \toprule
             ~ & \multicolumn{5}{c}{CIFAR-10} & \multicolumn{5}{c}{CIFAR-100} \\ 
                \cline{2-11}
            ~ & \textbf{PGD} & \textbf{BIM} & \textbf{FGSM} & \textbf{DeepFool} & \textbf{C\&W} & \textbf{PGD} & \textbf{BIM} & \textbf{FGSM} & \textbf{DeepFool} & \textbf{C\&W} \\
           \midrule
            learning rate & 1e-2 & 1e-2 & 1e-2 & 1e-2 & 1e-2 & 1e-2 & 1e-2 & 1e-2 & 1e-2 & 1e-2  \\ 
            momentum  & 0.9 & 0.9 & 0.9 & 0.9 & 0.9 & 0.9 & 0.9 & 0.9 & 0.9 & 0.9 \\
            weight decay & 5e-4 & 5e-3 & 5e-3 & 5e-3 & 5e-3 & 5e-4 & 5e-3 & 5e-3 & 5e-4 & 5e-3 \\
           \bottomrule
            
        \end{tabular}
        }
    }
\label{tab: Setting-PRD}
\end{table*}

\begin{table*}[t]
\centering
    \renewcommand\arraystretch{1.3}
    \centering
        \caption{\small{Comparison of detection performance on CIFAR-10 dataset under white-box and black-box settings.}}
        \resizebox{0.9\textwidth}{!}{%
        \setlength\tabcolsep{12pt}
        \scalebox{1.00}{
        \begin{tabular}{cccccccc}
            
            \toprule[1.5pt]
            
             \makebox[0.1\textwidth][c]{\multirow{3}{*}{\Large Dataset} }
            &\makebox[0.18\textwidth][c]{\multirow{3}{*}{\Large Method} }
            &\makebox[0.08\textwidth][c]{\makebox[0.08\textwidth][c]{\multirow{2}{*}{White-box}}}
            &\multicolumn{4}{c}{\makebox[0.08\textwidth][c]{\large BAD Evaluation}}\\ 
                \cline{4-7}
            ~ & ~ & ~ & SM-AT & SM-NT &  EM-AT & EM-NT \\
                \cline{3-7}
            ~ & ~ & FGSM/BIM/PGD-$l_{2}$ & FGSM/BIM/PGD-$l_{2}$ & FGSM/BIM/PGD-$l_{2}$ & FGSM/BIM/PGD-$l_{2}$ & FGSM/BIM/PGD-$l_{2}$ \\
                
                \midrule[1.0pt]
            
            \multirow{23}{*}{\Large CIFAR-10}
            & LID & 73.52/78.65/74.40 & 50.46/50.18/50.64 & 82.15/73.28/54.55 & 43.73/44.14/46.68 & 61.84/54.56/47.27\\
            
            & MD  & 79.43/89.05/89.66 & 57.06/63.22/63.70 & 87.57/81.37/74.84 & 40.22/54.12/53.89 & 78.31/68.10/65.59 \\
            
            & SID & 85.93/86.03/90.97 & 60.42/54.80/55.49 & 86.69/73.75/75.91 & 56.83/49.60/54.08 & 81.72/63.81/77.70 \\
            \cline{2-7}
            
            &    PRD-base & 87.29/90.96/95.48 & 62.75/68.76/69.54 & 88.84/84.04/94.10 & 46.97/41.14/56.65 & 83.40/81.82/90.04 \\
            
            &   PRD-online & 83.45/85.62/92.65 & 69.29/68.72/72.41 & 88.41/85.9791.26 & 62.88/54.43/75.87 & 86.14/81.25/94.49 \\
            \cline{2-7}
            
            &    FRD-base(joint) & 90.23/87.95/97.66 & 64.03/66.39/68.63 & 89.86/82.69/95.29 & 50.34/47.27/54.14 & 83.85/75.88/94.94 \\
            
            &    FRD-base(amp) & 91.54/87.94/97.68 & 63.54/66.43/68.63 & 90.57/82.69/95.34 & 49.47/47.21/54.16 & 85.72/75.86/94.84 \\ 
            
            &    FRD-base(pha) & 91.73/87.68/97.48 & 64.98/66.58/69.13 & 91.44/82.47/93.27 & 51.51/48.40/58.31 & 86.66/77.21/95.91 \\
            \cline{2-7}
            
            &   FRD-online(joint) & 88.47/84.29/97.21 & 73.08/68.20/74.43 & 91.34/84.25/95.02 & 67.26/58.81/70.20 & 90.05/82.35/98.62 \\
            
            &   FRD-online(amp) & 95.26/84.56/98.49 & 67.63/66.75/76.40 & 93.42/84.40/96.50 & 60.42/58.15/67.72 & 91.41/83.25/99.59 \\ 
            
            &   FRD-online(pha) & 90.32/79.98/96.24 & 70.78/66.65/76.27 & 91.14/82.60/95.36 & 63.27/56.08/70.93 & 88.99/74.80/99.37 \\ 
            
            \cline{2-7}
            \cline{2-7}
            
            ~ & ~ & PGD-$l_{\infty}$/DF/C\&W & PGD-$l_{\infty}$/DF/C\&W & PGD-$l_{\infty}$/DF/C\&W & PGD-$l_{\infty}$/DF/C\&W & PGD-$l_{\infty}$/DF/C\&W \\

            \cline{2-7}
            
            ~
            & LID & 72.03/80.99/80.93 & 51.43/72.27/68.93 & 51.27/80.85/85.05 & 47.71/64.59/58.49 & 46.87/75.33/73.75 \\
            
            & MD  & 86.07/92.23/89.58 & 60.26/62.66/55.53 & 67.53/82.78/87.82 & 52.70/50.88/38.53 & 61.78/71.13/79.71 \\
            
            & SID & 89.20/91.73/90.28 & 54.78/64.53/57.84 & 71.82/84.24/85.74 & 52.73/61.40/54.05 & 73.56/84.87/81.17 \\
            
            \cline{2-7}
            &    PRD-base & 94.25/96.72/85.83 & 72.07/72.58/57.79 & 89.01/92.96/84.69 & 60.33/64.16/41.27 & 90.69/88.05/76.79 \\
            &   PRD-online & 89.55/89.41/81.81 & 73.84/77.66/67.21 & 87.54/90.58/87.72 & 70.40/77.20/58.47 & 94.59/91.59/77.27 \\
            
            \cline{2-7}
            &    FRD-base(joint) & 95.41/96.93/83.38 & 70.66/74.38/59.54 & 91.99/94.13/85.37 & 57.41/65.23/43.76 & 92.58/89.67/76.97 \\ 
            
            &    FRD-base(amp) & 95.42/96.98/83.29 & 70.58/73.27/57.58 & 92.00/94.19/85.36 & 57.44/65.27/43.70 & 92.59/89.78/76.95 \\
            
            &    FRD-base(pha) & 95.11/97.04/83.73 & 71.91/72.25/58.29 & 90.17/95.80/83.81 & 62.17/65.46/44.41 & 93.25/91.26/75.83 \\
            \cline{2-7}
            
            &   FRD-online (joint) & 96.79/98.43/84.36 & 79.93/78.56/66.38 & 94.77/96.48/87.34 & 66.79/78.07/59.47 & 99.33/98.66/80.49 \\
            
            &   FRD-online (amp) & 96.55/98.38/84.23 & 79.71/79.02/66.28 & 94.51/96.46/87.28 & 68.41/78.21/59.88 & 99.27/98.53/80.90 \\
            
            &   FRD-online (pha) & 94.87/95.76/83.37 & 80.74/80.61/67.31 & 93.49/94.51/86.02 & 71.79/78.18/60.68 & 98.88/96.34/79.18 \\ 
            
            \bottomrule[1.5pt]
            \end{tabular}

    }}
\label{tab:comparisons_cifar10}
\end{table*}

\begin{table*}[b]
\centering
    \renewcommand\arraystretch{1.3}
    \centering
        \caption{\small{Comparison of detection performance on CIFAR-100 dataset under white-box and black-box settings.}}
        \resizebox{0.9\textwidth}{!}{%
        \setlength\tabcolsep{12pt}
        \scalebox{1.00}{
        \begin{tabular}{cccccccc}
            
            \toprule[1.5pt]
            
            \makebox[0.1\textwidth][c]{\multirow{3}{*}{\Large Dataset} }
            &\makebox[0.18\textwidth][c]{\multirow{3}{*}{\Large Method} }
            &\makebox[0.08\textwidth][c]{\makebox[0.08\textwidth][c]{\multirow{2}{*}{White-box}}}
            &\multicolumn{4}{c}{\makebox[0.08\textwidth][c]{\large BAD Evaluation}}\\ 
                \cline{4-7}
            ~ & ~ & ~ & SM-AT & SM-NT &  EM-AT & EM-NT \\
                \cline{3-7}
            ~ & ~ & FGSM/BIM/PGD-$l_{2}$ & FGSM/BIM/PGD-$l_{2}$ & FGSM/BIM/PGD-$l_{2}$ & FGSM/BIM/PGD-$l_{2}$ & FGSM/BIM/PGD-$l_{2}$ \\
                
                \midrule[1.0pt]
            
            \multirow{23}{*}{\Large CIFAR-100}
            & LID & 60.56/76.35/70.50 & 58.81/42.55/38.22 & 77.63/61.08/52.20 & 33.22/69.74/71.75 & 43.61/58.80/59.81 \\
            
            & MD  & 77.17/90.00/86.82 & 55.11/62.52/62.60 & 77.48/80.16/73.20 & 56.35/61.03/59.87 & 56.55/64.47/64.28 \\
            
            & SID & 67.73/89.11/88.65 & 65.54/63.45/66.96 & 86.99/80.41/76.51 & 47.82/74.42/81.04 & 62.91/76.18/84.11 \\
            
            \cline{2-7}
            
            &    PRD-base & 89.03/93.90/97.27 & 56.02/64.31/64.62 & 88.79/89.82/95.65 & 67.12/74.12/91.71 & 73.09/74.14/91.17 \\
            
            &   PRD-online & 95.12/92.17/99.38 & 57.40/64.20/74.31 & 92.21/89.97/97.76 & 60.77/60.00/90.67 & 66.13/70.36/92.90 \\
            \cline{2-7}
            
            &    FRD-base(joint) & 91.33/90.48/99.03 & 61.56/66.13/63.91 & 91.35/88.68/98.49 & 69.98/70.24/93.12 & 72.53/72.35/93.68 \\
            
            &    FRD-base(amp) & 91.35/90.48/98.96 & 61.49/66.14/63.30 & 91.32/88.65/98.38 & 75.68/72.75/93.19 & 77.98/74.54/94.33 \\ 
            
            &    FRD-base(pha) & 93.48/91.31/97.47 & 64.36/67.22/55.03 & 93.08/88.87/95.95 & 74.61/72.12/94.59 & 77.24/73.40/95.58 \\
            \cline{2-7}
            
            &   FRD-online(joint) & 92.46/90.33/98.96 & 57.41/65.69/74.05 & 91.79/88.51/98.39 & 64.99/61.51/98.29 & 69.02/66.33/98.54 \\
            
            &   FRD-online(amp) & 95.33/91.54/99.15 & 57.87/67.62/75.18 & 93.27/89.88/98.97 & 69.32/56.63/98.35 & 73.41/62.41/98.60 \\ 
            
            &   FRD-online(pha) & 94.81/91.22/98.33 & 58.79/66.94/69.83 & 92.25/90.10/96.83 & 68.71/62.04/98.41 & 71.22/65.34/98.75 \\ 
            
            \cline{2-7}
            \cline{2-7}
            ~ & ~ & PGD-$l_{\infty}$/DF/C\&W & PGD-$l_{\infty}$/DF/C\&W & PGD-$l_{\infty}$/DF/C\&W & PGD-$l_{\infty}$/DF/C\&W & PGD-$l_{\infty}$/DF/C\&W \\

            \cline{2-7}
            
            ~
            & LID & 63.83/67.90/66.05 & 40.48/59.77/55.61 & 43.50/77.87/78.04 & 71.81/38.15/36.49 & 60.05/45.82/44.64 \\
            
            & MD  & 73.01/86.02/80.90 & 58.78/58.15/56.95 & 57.31/83.24/78.42 & 65.78/59.79/63.74 & 66.77/60.66/65.50 \\
            
            & SID & 77.52/91.22/69.98 & 64.46/71.25/64.42 & 62.12/88.29/86.67 & 76.78/53.61/49.05 & 76.77/68.71/64.92 \\
            
            \cline{2-7}
            &    PRD-base & 90.83/88.57/85.87 & 77.53/59.78/58.18 & 88.35/89.09/88.22 & 93.15/67.97/61.06 & 89.97/69.57/62.61 \\
            
            &   PRD-online & 97.85/95.77/88.63 & 82.27/68.75/59.65 & 97.17/94.07/87.22 & 99.61/77.40/58.78 & 99.21/78.23/60.61 \\
            
            \cline{2-7}
            &    FRD-base(joint) & 88.68/96.69/86.39 & 77.22/70.18/59.89 & 85.84/95.73/86.44 & 98.43/81.35/60.28 & 98.14/84.45/64.13 \\ 
            
            &    FRD-base(amp) & 89.10/96.69/86.40 & 77.31/69.78/59.96 & 86.32/95.88/86.42 & 98.63/81.35/60.31 & 98.33/84.45/64.16 \\
            
            &    FRD-base(pha) & 81.86/96.07/86.38 & 76.90/63.90/60.47 & 78.33/95.29/86.43 & 97.77/78.69/60.89 & 97.61/81.72/63.78 \\
            \cline{2-7}
            
            &   FRD-online (joint) & 95.60/96.63/87.84 & 81.07/68.34/59.78 & 89.20/95.78/87.01 & 99.88/86.24/52.52 & 99.79/88.75/57.18\\
            
            &   FRD-online (amp) & 96.19/97.27/88.55 & 80.61/69.87/60.38 & 95.09/96.95/86.42 & 99.88/86.44/52.54 & 99.79/88.96/57.19 \\
            
            &   FRD-online (pha) & 94.21/96.90/89.30 & 79.92/69.59/60.55 & 92.86/97.00/86.96 & 99.69/86.55/53.27 & 99.62/89.43/60.05 \\ 
            
            \bottomrule[1.5pt]
            \end{tabular}

    }}
\label{tab:comparisons_cifar100}
\end{table*}

\end{document}